\documentclass[spanish]{IEEEtran}   	% use "amsart" instead of "article" for AMSLaTeX format
\usepackage{geometry}                		% See geometry.pdf to learn the layout options. There are lots.
\usepackage[english]{babel}
\usepackage[utf8]{inputenc}
\usepackage{graphicx}				% Use pdf, png, jpg, or eps§ with pdflatex; use eps in DVI mode
\usepackage[numbers,sort&compress]{natbib}
\usepackage{amssymb}
\usepackage{amsmath} 
\usepackage{hyperref}
\usepackage{authblk}
\usepackage{url,overcite}
\usepackage{tikz}
\usetikzlibrary{decorations.pathreplacing,calc,positioning,patterns}

\newcommand*{\mybox}[2]{\colorbox{#1!30}{\parbox{.98\linewidth}{#2}}}
\newcommand*{\coacts}{{\Omega}}
\newcommand*{\coactrv}{{Y}}

\newcommand*{\lp}{\underline{E}}
\newcommand*{\up}{\overline{E}}
\usepackage{mathtools}
  \mathtoolsset{mathic}
    \allowdisplaybreaks

\begin{document}

\title{Generalisation of Total Uncertainty in AI: A Theoretical Study}
\author[1]{Keivan Shariatmadar}

\affil[1]{Mecha(tro)nic System Dynamics (LMSD), Bruges Campus, KU Leuven, Belgium}
\renewcommand\Authands{, }
\date{}							% Activate to display a given date or no date

\maketitle

\textit{}\textbf{Abstract}
AI has been dealing with uncertainty to have highly accurate results. This becomes even worse with reasonably small data sets or a variation in the data sets. This has far-reaching effects on decision-making, forecasting and learning mechanisms. This study seeks to unpack the nature of uncertainty that exists within AI by drawing ideas from established works, the latest developments and practical applications and provide a novel total uncertainty definition in AI. 

From inception theories up to current methodologies, this paper provides an integrated view of dealing with better total uncertainty as well as complexities of uncertainty in AI that help us understand its meaning and value across different domains.

\section{\textbf{Introduction}}

Uncertainty is one of the greatest determinants of AI research landscapes and applications. The significance of uncertainty as an inevitable element to consider when unlocking the full potential of artificial intelligence technologies grows exponentially with advances made within this field across various sectors. In other words, uncertainty is a characteristic of the real world. It touches on all aspects of human life. The existence of artificial intelligence (AI) relies on the ability to understand and manage uncertainty effectively in order to build dependable, resilient, and adaptive (intelligent) systems. There are different forms in which uncertainty appears: from incomplete or noisy data to situations that can have multiple outcomes at once. Uncertainty must be addressed not as an abstract concept but as a practical necessity since AI systems often function within volatile and ambiguous conditions where choices have to be made based on imperfect information.

Furthermore, the importance of uncertainty in AI cannot be underestimated. In other words, AI seeks human-like intelligence meaning it is reasoning, inferring, and making decisions even when there is no complete certainty involved. For example, autonomous vehicles could negotiate with unpredictable traffic jams while medical diagnosis programs may interpret vague symptoms to suggest possible treatments for them. In these scenarios, AI models have to deal with uncertainties before they can come up with correct and well-founded judgments that are also safe enough \citep{Lemmer1988-LEMUIA}.

Additionally, uncertainty characterises every phase in the entire pipeline of AI including data preprocessing; feature engineering; model training/testing/evaluation/deployment.

Besides, uncertainty is all over AI pipeline stages involving data preprocessing and feature engineering, model training and analysis and deployment. In AI approaches that are data-driven, uncertainty comes from inherent uncertainty in the data, sampling variability, imperfect models, or model approximation errors. Even in rule-based or symbolic AI systems, however, any conclusions will be affected by uncertainty based on the complexity of real-world phenomena as well as human reasoning. This paper brings together succinctly important phenomena, approaches, as well as new directions taken by AI systems to deal with uncertainty as they are found in Kanal \citep{Lemmer1988-LEMUIA}.

This paper is motivated by a growing understanding of the new definition of total uncertainty defined by two types of uncertainties in machine learning---Epistemic and Aleatoric uncertainties, explained in Section \ref{sec:EaAU}. 

The objectives of this paper include unravelling the multifaceted nature of uncertainty in artificial intelligence and exploring its theoretical bases, practical implications as well as emerging trends with a better representation of Epistemic and Aleatoric uncertainties. Moreover, this paper provides a comprehensive and insightful exploration of AI uncertainties by synthesising insights from seminal works, recent advancements and practical applications which can contribute towards better understanding for researchers, practitioners and enthusiasts. By compiling theoretical perspectives, empirical methodologies and future applications, this work can be said to encompass the full range of uncertainty in artificial intelligence. It serves scientific purposes by benchmarking the state-of-the-art information from several important sources with the aim of enlightening scholars, professionals and interested individuals on various dimensions of AI uncertainties versus the development of intelligent systems.

In subsequent sections, we will take a ride to see how uncertainly can be chaotic in AI. We will look at foundational uncertainty definitions and related problems in Section \ref{sec:EaAU}. Different methods to estimate imprecise uncertainty are explained in Section \ref{sec:UEst}. Conclusion and future works are discussed in Section \ref{sec:Conc}.

\subsection{\textbf{Epistemic and aleatoric uncertainties}}
The importance of understanding and quantifying uncertainty in machine learning has increased
significantly. A proper understanding of the uncertainty framework that defines the types of
optimisation problems is important and we encounter them in this paper. Therefore, a proper
delimiting of the scope as well as understanding of the underlying ideas of the selected concepts
and their impact on the loss function to be optimised is necessary.
This is primarily due to the growing complexity of AI applications and the rapid growth of data. The
need to address this demand arises from the unpredictable nature of utilising AI in diverse domains,
including safety-critical areas like healthcare and autonomous systems, to more business-related
such as finance and marketing. Machine learning has the potential to revolutionise various industries.
However, it also poses inherent risks, particularly in terms of generalisation, domain adaptation, and
making safe decisions in many applications, e.g., control and decision making. To tackle this problem, researchers have been continuously developing uncertainty quantification methods in AI. Unlike standard machine learning algorithms, an advanced uncertainty-equipped machine learning algorithm provides us with a set of predictions for possible outputs instead of a single point prediction. This ability to return a set of possible outcomes helps us to make better and safer decisions in different scenarios.
In this section, we primarily give an overview of different methods and algorithms that enable
machine learning methods to quantify and approximate uncertainty. Several methods that we cover
in this section, include direct interval prediction, ensemble models, Bayesian methods, Random
sets \& Belief function models, and conformal prediction. We mainly discuss these four methods
due to their popularity.

\subsubsection*{\textbf{Aleatoric uncertainty}}
Aleatoric (statistical) uncertainty (AU) refers to randomness or variability: What a random sample drawn from a probability distribution will be.

\subsubsection*{\textbf{Epistemic uncertainty}}
 Epistemic (systematic) uncertainty (EU) refers to the lack of knowledge: What is the relevant probability distribution?

\subsection{\textbf{Total uncertainty}}\label{sec:total}
In machine learning, the total uncertainty (TU) is defined by \citep{ABELLAN2005235},
\begin{equation}\label{eq-total}
    \mathbf{TU = EU + AU}.
\end{equation}
The definition \eqref{eq-total} is correct when the epistemic and aleatoric uncertainties are independent. As a simple example, the mean of two entities could be different with different noise levels i.e., EU and AU are dependent and we need a better definition of total uncertainty. In the next Section \ref{sec:UEst} we explain a new definition for TU and an overview of advanced (imprecise) uncertainty models.
\section{Uncertainty models and Identifications}\label{sec:EaAU}
In this section, we focus on an overview of the analytical concepts of some advanced uncertainty models that we use in WP2, e.g., intervals, random sets, probability intervals, and credal sets. We explain the essential ideas behind the concepts. These advanced models are set models and deal with in-determinism in the uncertainty (second-order uncertainty). 

In general, and specifically in engineering and machine learning domains, uncertainty is defined in two categories: aleatoric and epistemic uncertainty. Aleatoric uncertainty arises from the inherent randomness in the data, and it is irreducible. In contrast, epistemic uncertainty arises from the lack of information or data which is reducible (by gathering more data). 

In this section, our focus is on integrating epistemic uncertainty models into the optimisation problems under uncertainty. The goal is to investigate different types of epistemic uncertainty models for the optimisation problem under uncertainty that enables ``pointwise'' optimisation under epistemic uncertainty. While robust and stochastic optimisation methods are commonly employed in dealing with optimisation under the Bayesian uncertainty, they fail to account for predicting epistemic uncertainty with acceptable accuracy. By incorporating epistemic uncertainty models, we can enhance the optimisation process by accounting for the lack of information or data. 
To model the epistemic uncertainty, we propose several epistemic uncertainty models such as the (probabilistic) interval model, random set, and credal set models. 

\subsubsection{\textbf{Deterministic Intervals}}
The deterministic interval model—also called vacuous previsions— expresses ignorance relative to a non-empty subset of possible parameter values. The interval model is the least informative model. The interval representation only needs the lower and upper values of an uncertain parameter. Hence, the knowledge of the probabilistic distribution over the parameters between its lower and upper value is not required. Mathematically, an interval model
is written as $X = [\underline{X}, \overline{X}]$.

The interval model, also known as vacuous expectation, represents ignorance regarding a nonempty subset of possible parameter values. In state-of-the-art problems, the Interval arithmetic attempts to propagate the interval through the problem to find a resulting interval, aiming to calculate the upper and lower bounds of this propagated interval. However, precise calculation of these bounds can be challenging or even impossible, and the complexity of this method for higher-dimensional e.g., optimisation problems is NP-hard \citep{ROKNE199261, Oliveira-Antunes-2007}. 

The lower and upper expectations for a subset \( A \) of \( \coacts \) and for a given functional \( g := g(Y) \) of the random variable \( Y : A \mapsto \mathbb{R} \) are defined as follows:\\
\mybox{darkgray}{Interval Model}
\mybox{lightgray}{
\begin{equation}\label{eq:intmooo}
\lp(g) \coloneqq \inf g\vert_A \quad \text{and} \quad \up(g) \coloneqq \sup g\vert_A.
\end{equation}
}\\
When \( A := [a,b] \subset \Omega := \mathbb{R} \) (one-dimensional case) and \( g(y) \) is a continuous function on \( y \in [a,b] \), the lower and upper expectations are defined as:\\
\mybox{lightgray}{
\begin{equation}\label{eq:intpree}
\lp(g) = \min_{y \in [a,b]} g(y) \quad \text{and} \quad \up(g) = \max_{y \in [a,b]} g(y).
\end{equation}
}\\
This model effectively represents the available information when it is known that \( \coactrv \) assumes values in \( A \) but no further probabilistic information is available, other than \( P(A) = 1 \).

\subsubsection{\textbf{Probability Intervals}}
In classification of $C$ elements, probability intervals can be defined
as $[\underline{\boldsymbol{y}}, \overline{\boldsymbol{y}}]\!=\!\{[\underline{y}_k, \overline{y}_k]\}_{k=1}^C$
, where $0\leq\underline{y}_k\leq\overline{y}_k\leq1$ . They represent the lower and upper bounds of the probabilities associated with the relevant classes.

\subsubsection{\textbf{$\epsilon$-contaminations}}
\label{epsisec}
 This model is more advanced than the two previous models---interval and probability intervals. It is easy to build as well as implement compared to the other uncertainty models such as a possibility distribution model. The contamination model is a mixed/hybrid model, which is built by two simple models---a non-probabilistic model such as interval and a probabilistic model such as a probability distribution. The state-of-the-art of the mixed models\footnote{To a Bayesian analyst (a researcher who works on the deterministic uncertainty framework), the distinction between fixed, random and mixed models boils down to a specification of the number of stages in a given hierarchical model. In literature, several classes of prior distribution have been proposed but the most commonly used one is the contamination class, e.g., works of Good \citep{good1965estimation}, Huber \citep{huber1973}, Dempster \citep{DEMPSTER1977121}, Rubin \citep{RUBIN1977351}, and Berger \citep{1985:berger} to mention a few.} is mainly focused on Bayesian sensitivity analysis and 
not much attention has been paid to non-deterministic advanced uncertainty cases (such as intervals or high-dimensional cases like probability boxes). 

As mentioned before, when there are insufficient data available to build a probabilistic model we can start with the interval model (the least informative model). While working with the interval model, more data can be obtained via, e.g., extra tests, further analysis, access to other similar resources/data, and so on. Therefore, a contamination model can be generated that considers both models. This model links/contaminates the precise model (a distribution) with an imprecise model (an interval). This is one of the important properties of this model to identify imprecision in a given uncertainty model. We explained this in another paper \citep{keivan2021:ICUME}.  
We develop a mixed model for a given probability measure $P$ and a lower expectation $\underline E_I$, given via an interval (imprecise) model.
\subsection*{Definition}\label{epsi:def}
 An $\epsilon$-contamination model $\underline E(\cdot)$ is described as a convex combination of two uncertainty models: (i) Probabilistic/precise model, e.g., Normal distribution with expectation $E$, (ii) non-probabilistic/imprecise model, e.g., interval model with lower expectation $\underline E_I$, defined in \eqref{eq:intpree}. For a gamble $f\in\mathcal{G}(\Omega)$ the lower expectation is described as follows:\\
\mybox{darkgray}{Contamination model}\\
\mybox{lightgray}{
\begin{equation}\label{lpepsi}
\underline E(f)=(1-\epsilon)E(f)+\epsilon\underline E_I(f)
\end{equation}
}\\
where \(E\in\mathcal{M}\left\{\underline E_I\right\}=\left\{E:\forall f\in\mathcal{G}(\Omega),E(f)\ge\underline E_I(f) \right\}\) is the set of dominating linear expectations by $\underline E_I$. By definition \eqref{eq:intmooo}, the lower and upper expectation for a given interval $[a,b]$ is defined as
\begin{equation}\label{loepsi}
\underline E_I(f(y))=\inf_{y\in[a,b]}f(y)~~\text{and}~~\overline E_I(f(y))=\sup_{y\in[a,b]}f(y).
\end{equation}
\(0<\epsilon<1\) is called (in this dissertation) a tuning parameter or level of {model-trust/importance}. Similarly, the upper expectation is\\
\mybox{lightgray}{
\begin{equation}\label{upepsi}
\overline E(f)=(1-\epsilon)E(f)+\epsilon\overline E_I(f).
\end{equation}
}

\subsubsection{\textbf{Credal Sets}}
\begin{figure}[!htbp]
\begin{center}
\includegraphics[width=4.7cm]{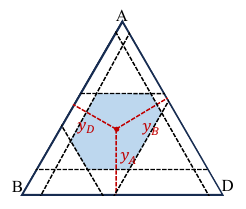}
\caption{Convex-closed set of probability distributions (credal set, in blue).} 
\label{FIG: Credal}
\end{center}
\end{figure}

A credal set is defined as a convex set of probability distributions. One of the computationally effective ways to construct a credal set, denoted as $\mathbb{Q}$, is to use probability intervals, as follows:
\begin{equation*}
\mathbb{Q} \!=\! \{\boldsymbol{y=(y_1,\cdots,y_C)}\!\mid\! y_k \!\in\! [\underline{y}_k, \overline{y}_k],
\end{equation*}
\begin{equation}
\forall k \!=\!1, 2, ..., C  \},
\label{Eq: CredalPIs}
\end{equation}
in which $\boldsymbol{y}$ represents a single probability distribution vector. Here, $\mathbb{Q}$ is a special convex set (a
polytope) of probabilities constrained by probability intervals.
To prevent $\mathbb{Q}$ from being empty, $[\underline{\boldsymbol{y}}, \overline{\boldsymbol{y}}]$ is required to satisfy the following condition \citep{probability_interval_1994}:
\begin{equation}
\textstyle\sum\nolimits_{k=1}^{C} \underline{y}_k \leq 1 \leq  \sum\nolimits_{k=1}^{C} \overline{y}_k.
\label{Eq: proper_p_interval}
\end{equation}
Figure \ref{FIG: Credal} illustrates generating a credal set from probability intervals for a three-class classification problem. It is identified on the probability simplex via probability intervals for three classes, a triangle of all probability distributions on $\mathbb{T}\!=\!\{\text{A}, \text{B}, \text{D}\}$. In a probability simplex, a discrete probability distribution over classes $(y_A, y_B, y_D)$ is represented as a single point. The distance between each point and the edge opposite the vertex representing a class is the probability assigned to that class (red dashed lines). The probability intervals act as constraints (parallel black dashed lines) which determine the credal set.

When a neural network prediction assumes the form of a credal set $\mathbb{Q}$ (CredNN), an extension of Shannon entropy has been proposed to measure the uncertainty \citep{hartley-mes},
\begin{equation}\label{eq:shan}
H^*\!=\! \max_{\boldsymbol{y}\in\mathbb{Q}}\!H(\boldsymbol{y}), H_* \!=\! \min_{\boldsymbol{y}\in\mathbb{Q}}\!H(\boldsymbol{y}),
\end{equation}
in which $H^*$ and $H_*$ denote the upper and lower Shannon entropy, and serve as the measure for the total (TU) and aleatoric uncertainty (AU), respectively. Epistemic uncertainty (EU) is then measured by the difference $H^{*} \!-\! H_{*}$.

\subsubsection{\textbf{Random Sets}}
\label{Subsec: RandomSet}
In a traditional classifier like a neural network, the output is a mapping from input data to a single category. However, in set-valued classification, the output is a mapping from the input data to a set of possible categories. For instance, instead of predicting a single class, a set-valued classifier might predict multiple potential classes for a given input.
In set-valued classification, the prediction for each input is not a vector of softmax probabilities as in traditional classification. Instead, it's a belief function, where each output corresponds to a subset of possible classes. For example, if there are $N$ classes, a basic set-valued classifier would have $2^N$ outputs, each representing a different combination of classes.
To overcome the exponential complexity of using $2^N$ sets of classes (especially for large $N$), a fixed budget of $K$ relevant non-singleton (of $cardinality~> 1$) focal sets are used. These focal sets are obtained by clustering the original classes and selecting the top $K$ sets of classes with the highest overlap ratio, computed as the intersection over union for each subset. The clustering is performed on feature vectors of images of each class generated by a standard CNN trained on the original classes.
The feature vectors are further reduced to 3 dimensions using t-SNE (t-Distributed Stochastic Neighbour Embedding) before applying a Gaussian Mixture Model (GMM) to them. Ellipsoids, covering $95\%$ of data, are generated using eigenvectors and eigenvalues of the covariance matrix and the mean vector obtained from the GMM to calculate the overlaps. To avoid computing a degree of overlap for all subsets, the algorithm is early stopped when increasing the cardinality does not alter the list of most overlapping sets of classes. The non-singleton focal sets so obtained, along with the $2^N$ original (singleton) classes, form our network outputs E.g., in a $100$-class scenario, the power set contains subsets ($1030$ possibilities). Setting a budget of $K = 200$, for instance, results in $100 + K  = 300$ outputs, a far more manageable number.

\subsubsection{\textbf{Probability Box}}\label{fre:sec}
Another advanced and highly informative uncertainty model is the (generalised) \emph{probability box} or \emph{p-box}. When multiple cumulative distribution functions (CDFs) from a database or experiment are available but a true distribution cannot be determined, they can be collected into a bounded set known as a p-box. In real-life scenarios, such as those involving numerous disturbances and a rapidly changing environment, finding a true distribution model can be challenging. In such cases, modelling uncertainty via a single unique distribution is impractical. However, with sufficient data, lower and upper distribution functions can be defined, within which any estimated distribution must lie.

\begin{figure}[h!]
  \centering
  \includegraphics [scale=0.35]{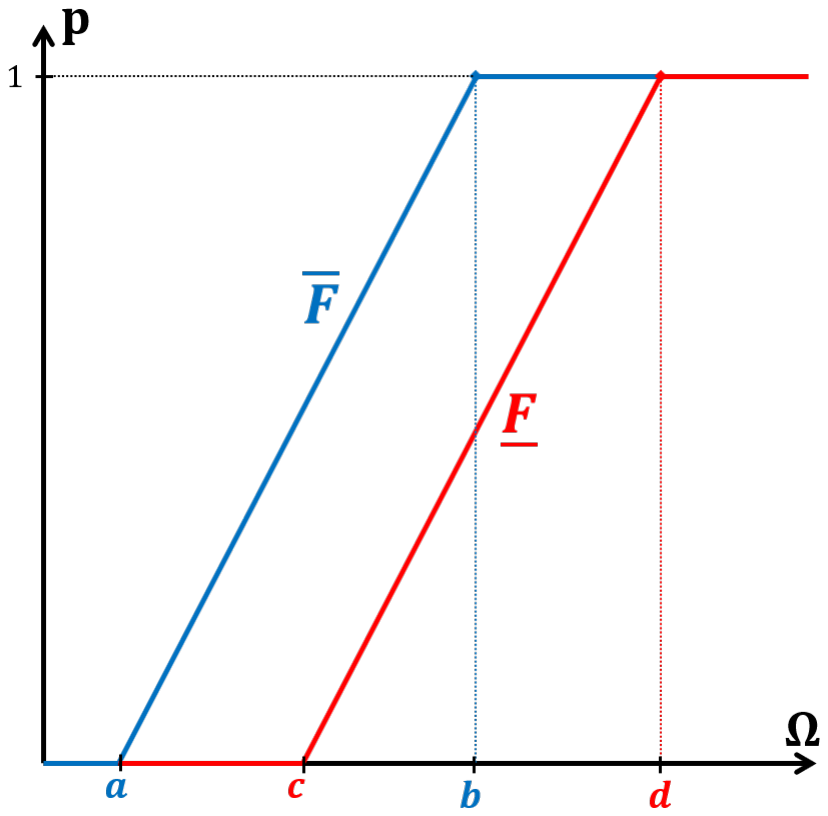}
  \caption{$\overline{F}:=U(a,b)$ is the upper bound and $\underline{F}=U(c,d)$ is the lower bound with uniform distributions}
  \label{fig:pbox}
\end{figure}

A p-box is a set of all distributions bounded by the lower and upper distribution functions. Mathematically, a (generalised) p-box is defined as a set of distributions bounded by a pair $(\underline{F}, \overline{F})$ of cumulative distribution functions mapping the sample space $\Omega$ to $[0, 1]$, satisfying $\underline{F} \leq \overline{F}$. If $\Omega$ is a closed interval on $\mathbb{R}$, the pair $(\underline{F}, \overline{F})$ is referred to as a p-box. Figure \ref{fig:pbox} illustrates a simple p-box with uniform distribution functions as the lower and upper bounds.

The CDF is any non-decreasing function $F:\Omega \longrightarrow [0, 1]$ that satisfies $F(1_{\Omega}) = 1$, where $1_{\Omega}$ is the largest member of $\Omega$ and $0_{\Omega}$ is the smallest. $F(s)$ provides information about the cumulative probability on the interval $[0_{\Omega}, s]$\footnote{Cumulative distribution functions are not assumed to be right-continuous.}. Given a cumulative distribution $F$ on $\Omega$ and a value $s \in \Omega$, $F(s^+)$ is the \emph{right-limit} and $F(s^-)$ is the \emph{left-limit}, defined as follows:
\[ F(s^+) = \inf_{y > s} F(y) = \lim_{y \rightarrow s, y > s} F(y),\] 
\[F(s^-) = \sup_{y < s} F(y) = \lim_{y \rightarrow s, y < s} F(y). \]
In Walley's framework \citep{Walley-1991}, a generalised p-box is interpreted as a lower expectation (actually a lower probability) $\underline{E}_{\underline{F}, \overline{F}}$ on the set of events \[\mathcal{Z} \subset \Omega, \mathcal{Z} := [0_{\Omega}, 1_{\Omega}] = \left\{ [0_{\Omega}, s] : s \in \Omega \right\}\] 
\[\cup \left\{ (r, 1_{\Omega}] : r \in \Omega \right\}\] as follows:
\[ \underline{E}_{\underline{F}, \overline{F}}([0_{\Omega}, s]) := \underline{F}(s),\] 
\[\underline{E}_{\underline{F},
\ \overline{F}}((r, 1_{\Omega}]) := 1 - \overline{F}(r). \]

A generalised p-box is thus a set of cumulative distribution functions that lie between $\underline{F}$ and $\overline{F}$,
\begin{equation}\label{pboxCred}
\Phi(\underline{F}, \overline{F}) := \{ F: \underline{F} \leq F \leq \overline{F} \}.
\end{equation} 

The lower expectation $\underline{E}_{\underline{F}, \overline{F}}$ for a p-box $\Phi(\underline{F}, \overline{F})$ is defined as the lower envelope of $\lp_F$ where $F$ is between $\underline{F}$ and $\overline{F}$:\\
\mybox{lightgray}{
\begin{equation}\label{naturalPbox}
\underline{E}_{\underline{F}, \overline{F}}(f) = \inf_{F \in \Phi(\underline{F}, \overline{F})} \underline{E}_F(f)
\end{equation}}

for all gambles, $f$ on $\Omega$. 

For an event of the type \( A = [x_0 := 0_{\Omega}, x_1] \cup (x_2, x_3] \cup \dots \cup (x_{2n}, x_{2n+1}] \) for \( x_1 < x_2 < \dots < x_{2n+1} \) in $\Omega$, the lower expectation is defined as:
\begin{equation*}
\underline{E}_{\underline{F}, \overline{F}}(A) = \underline{F}(x_1) + \sum_{k = 1}^{n} \max\{0, \underline{F}(x_{2k+1}) - \overline{F}(x_{2k}) \}.
\end{equation*}
For more details, see \citep{Walley-1991}, Section 4.6.6 and \citep{Matt2005Thesis}, p. 93.

\subsection*{Limit Approximation of P-box}

Consider a p-box $(\underline{F}, \overline{F})$ on $\Omega$. Let $\{\underline{F}_n\}_n, \{\overline{F}_n\}_n$ be increasing and decreasing sequences of CDFs converging point-wise to $\underline{F}$ and $\overline{F}$, respectively. Assume $\underline{E}_n$ is the lower probability associated with $(\underline{F}_n, \overline{F}_n)$. Since $\underline{F}_n \leq \underline{F}$ and $\overline{F}_n \geq \overline{F}$, it follows that $\Phi(\underline{F}, \overline{F}) \subseteq \Phi(\underline{F}_n, \overline{F}_n)$ and Equation \eqref{naturalPbox} implies that $\underline{E}_n \leq \underline{E}$. Furthermore, $\underline{E}_n \leq \underline{E}_{n+1}$ for any $n \in \mathbb{N}$, so $\lim_n \underline{E}_n = \sup_n \underline{E}_n \leq \underline{E}$. Thus, the natural extension can be approximated as follows: $\underline{E}(f) = \lim_n E_n(f)$ for any gamble $f$.

\begin{figure}[h!]
  \centering
  \includegraphics [scale=0.3]{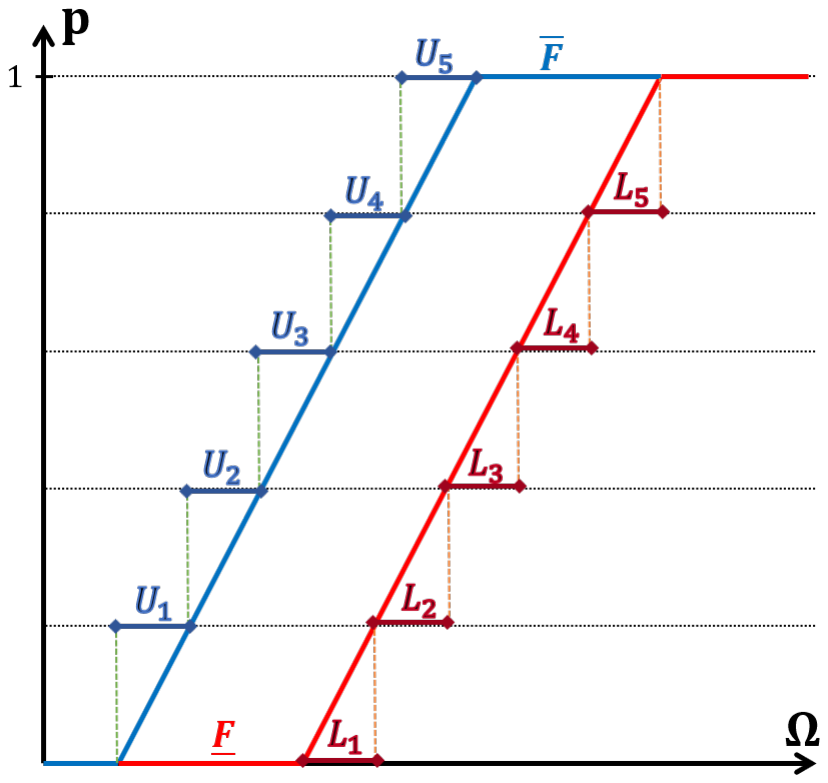}
  \includegraphics [scale=0.3]{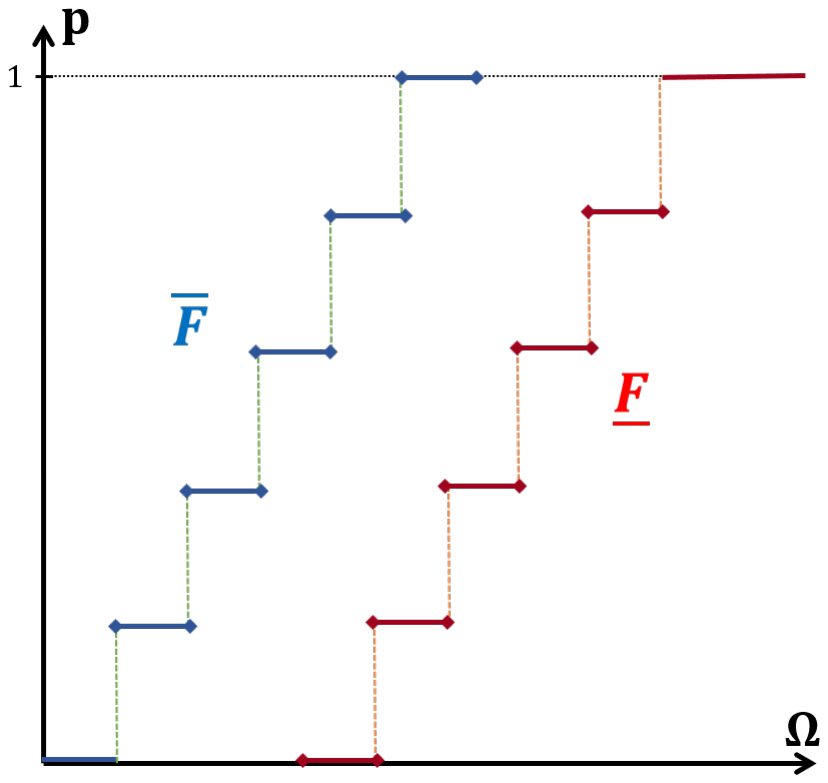}
  \caption{The blue $U_i$ lines represent the discrete upper bound for $\overline{F}$, and the red $L_i$ lines represent the discrete lower bound for $\underline{F}$ for $N=5$ partitions}
  \label{fig:dpbox}
\end{figure}

For a given p-box model---illustrated in Figure \ref{fig:dpbox} with the lower and upper bounds given by uniform distribution functions $\underline{F}$ and $\overline{F}$---we can discretise the $y$-axes (probability) into $n$ partitions, with $\underline{F}_n$ and $\overline{F}_n$ determined as follows:
\begin{equation}\label{app:pbox}
\underline{F}_n(y) = \begin{cases} 
\frac{i-1}{n} & \text{if } y \in L^n_i, y \neq 1_{\Omega}, \\
1 & \text{if } y = 1_{\Omega}
\end{cases}
\end{equation}
\begin{equation*}
\overline{F}_n(y) = \frac{i}{n} \text{ if } y \in U^n_i
\end{equation*}
where $n \in \mathbb{N}, i \in \{1, 2, 3, \ldots, n\}$ such that
\begin{align*}
U^n_i & := \overline{F}^{-1}\big([\frac{i-1}{n}, \frac{i}{n}]\big), \\
L^n_i & := \underline{F}^{-1}\big([\frac{i-1}{n}, \frac{i}{n}]\big).
\end{align*}

As shown, any p-box model can be approximated via $2n$ intervals ($U_i$'s and $L_i$'s). From the definition of the lower expectation under a \emph{discrete generalised p-box} \citep{keivan2019pbox, MIRANDA20151}, the approximations of the lower and upper expectations $\lp: \mathcal{G}(\Omega) \mapsto \mathbb{R}, \up: \mathcal{G}(\Omega) \mapsto \mathbb{R}$ for a gamble $f \in \mathcal{G}(\Omega)$ are given by:\\
\mybox{lightgray}{
\begin{equation*}
\underline{E}(f) = \lim_{n\rightarrow \infty} \left[\frac{1}{n} \sum_{i=1}^{n} \inf_{y \in L_i} f(y)\right]
\end{equation*}
\begin{equation}\label{app:dpbox}
\approx \frac{1}{N} \sum_{i=1}^{N} \inf_{y \in L_i} f(y),
\end{equation}
\begin{equation*}
\up(f) = \lim_{n \rightarrow \infty} \left[\frac{1}{n} \sum_{i=1}^{n} \sup_{y \in U_i} f(y)\right] 
\end{equation*}
\begin{equation}\label{eq:updpbox}
\approx \frac{1}{N} \sum_{i=1}^{N} \sup_{y \in U_i} f(y).
\end{equation}
}\\
where $L_i$ and $U_i$ are partitions of $\Omega := \mathbb{R}$, and $N$ is a (large) number of partitions. 

\section{Generalised Total Uncertainty Estimation}\label{sec:UEst}
As it is discussed in Section \ref{sec:total}, the definition \eqref{eq-total} is correct when AU and EU are independent. Here we will show via a simple (Counter) example that the
epistemic and aleatoric uncertainty are not necessarily independent.
\subsection{Dependency}
The mean of two the same data sets with different noises are different. Assume $X_1:=f(x)+\epsilon_1$ and $X_2:=f(x)+\epsilon_2$, where $\epsilon_1\sim N(\mu_1,\sigma_1)$ and $\epsilon_2\sim N(\mu_2,\sigma_2)$, therefore $E(X_1)\not=E(X_2)$. Similarly, assume $X_1=X_2$ are two exactly the same datasets. If we remove some $K$-number of data points for instance from $X_2$---meaning increasing the epistemic uncertainty in $X_2$---then the noise level ($\sigma_2$) in $X_2$ will be different than $\sigma_1$. Therefore, AU and EU are dependent.

\subsection{Proposal I}
We propose a new definition by a linear combination of the AU and EU as follows.
\vspace{4mm}\\
\mybox{lightgray}{
\begin{equation}\label{eq:propsI}
TU:=\alpha_1 AU + \alpha_2 EU,    
\end{equation}
}\vspace{4mm}\\
Since, by definition, the TU must be greater than either AU or EU, then $\alpha_1 + \alpha_2>1$. But how to define $\alpha_1$ and $\alpha_2$? Some preliminary ideas are proposed as follows. 
\begin{itemize}
    \item[1.] Investigate if the model is robust against increased noise or decreased data size. To check the sensitivity of the model to AU or EU. Then we make $\alpha_1$ smaller or greater than $\alpha_2$. 
\item[2.] In the case of CredNN, calculate the upper, $H^*$, and lower, $H_*$, Shannon entropy defined in \eqref{eq:shan}. The difference (\emph{imprecision}) between upper and lower Shannon entropy is the lower bound for EU, i.e., $\alpha_2\ge H^*-H_*$.
\item[3.] In the case of Interval Neural Network (INN) \citep{wang2024CreINN}, the difference between the upper, $U_{EU}$, and lower, $L_{EU}$, is the lower bound for EU, i.e., $\alpha_2\ge U_{EU}-L_{EU}$. 
\item[4.] In the case of deep ensembles (EnNN) \citep{wang2024credalwrap}, take the difference between highest, $U_{EnNN}$, and lowest, $L_{EnNN}$, prediction as $\alpha_2$, i.e., $\alpha_2\ge U_{EnNN}-L_{EnNN}$.
\end{itemize}

\subsection{Proposal II}
In the case of the contamination model \eqref{loepsi}, the precise part defines the AU and the imprecise part defines the EU. We call this model Contamination Neural Network (ContNN) which is defined as follows.
\vspace{4mm}\\
\mybox{lightgray}{
\begin{equation}\label{eq:contnn}
\text{ContNN} := \epsilon.\text{BNN} + (1-\epsilon).\text{INN}
\end{equation}
}\vspace{4mm}\\
where BNN is the Bayesian Neural Network \citep{BNNKaiz}.

We use BNN for the AU definition and INN for the EU definition of the ContNN novel model in \eqref{eq:contnn} as $AU_{ContNN}$ and $EU_{ContNN}$. Finally, we define the total uncertainty for the ContNN as, $TU_{ContNN}:=AU_{ContNN} + EU_{ContNN}$.

\section{\textbf{Conclusion and future work}}\label{sec:Conc}
In this paper, we proposed two novel ideas for the Total Uncertainty definition as proposals (I) and (II). In proposal (I), we provide four approaches to identify the parameters $\alpha_1$ and $\alpha_2$ in \eqref{eq-total}. The validation, comparison, and application will be discussed in our next paper. In proposal (II), we define a novel Neural Network via the convex combination of BNN and INN in \eqref{eq:contnn}. We define the novel Total Uncertainty, $TU_{ContNN}$, for the contamination neural network model. Based on the literature about BNN and INN (as well as CreNN), we have seen that EU could be better estimated with nob-Bayesian models e.g., INN. Furthermore, AU could be estimated better via the standard neural network or BNN. Therefore, the total uncertainty estimation for the novel ContNN is better defined compared to the definition \eqref{eq-total}. However, the ContNN model complexity is higher than the state-of-the-art models. In our future work, we will compare its pros and cons. 

\section{\textbf{Bibliography}}
\bibliographystyle{IEEEtran} 
\bibliography{references}

\end{document}